\documentclass[letterpaper]{article}
\usepackage{aaai19}
\usepackage{times}
\usepackage{helvet}
\usepackage{courier}
\usepackage{dsfont}
\usepackage{diagbox}
\usepackage{threeparttable}
\usepackage{xcolor, colortbl}
\usepackage{multirow}
\usepackage{graphicx}
\usepackage{booktabs}
\usepackage{enumerate,xspace}
\usepackage{url}
\usepackage{subcaption}
\usepackage{tikz}
\usepackage{mathtools}
\usepackage{amsmath}
\usepackage{amssymb}
\usepackage{tabularx}
\usepackage[]{floatrow} 
\usepackage{pgfplots}
\usetikzlibrary{positioning}
\usetikzlibrary{calc}
\usetikzlibrary{fit}
\usetikzlibrary{decorations.text}
\newcommand{\figref}[2][]{Figure#1~\ref{#2}\xspace}
\newcommand{\tabref}[2][]{Table#1~\ref{#2}\xspace}

\newcommand{\class}[1]{\textsf{#1}\xspace}
\newcommand{\FA}{\class{FA}}
\newcommand{\GA}{\class{GA}}
\newcommand{\Bart}{\class{B}}
\newcommand{\Cart}{\class{C}}
\newcommand{\Start}{\class{Start}}
\newcommand{\Stub}{\class{Stub}}

\newcommand{\lex}[1]{\textit{#1}\xspace}

\newcommand{\method}[1]{\textsc{#1}\xspace}
\newcommand{\inception}{\method{Inception}}
\newcommand{\bilstm}{\method{biLSTM}}
\newcommand{\joint}{\method{Joint}}
\newcommand{\majority}{\method{Majority}}
\newcommand{\benchmark}{\method{Benchmark}}
\newcommand{\doctovec}{\method{Doc2Vec}}
\newcommand{\frozen}{\method{Inception$_{\text{fixed}}$}}

\newcommand{\dataset}[1]{\textsf{#1}\xspace}
\newcommand{\wikipedia}{\dataset{Wikipedia}}
\newcommand{\arxiv}{\dataset{arXiv}}
\newcommand{\csai}{\dataset{cs.ai}}
\newcommand{\cslg}{\dataset{cs.lg}}
\newcommand{\cscl}{\dataset{cs.cl}}

\newcommand{\z}{\phantom{0}}
\newcommand{\ssig}{\ensuremath{\dagger}}
\newcommand{\nosig}{\phantom{\ensuremath{\dagger}}}

\newcommand{\precision}{\ensuremath{\mathcal{P}}}
\newcommand{\recall}{\ensuremath{\mathcal{R}}}
\newcommand{\fscore}{\ensuremath{\mathcal{F}_{\beta=1}}}

\newcommand{\citet}[1]{\citeauthor{#1}~\shortcite{#1}}

\begin{document}
	\title{A Joint Model for Multimodal Document Quality Assessment}
		\author{Aili Shen, Bahar Salehi, Timothy Baldwin, Jianzhong Qi\\
			School of Computing and Information Systems, The University of Melbourne, Victoria, Australia\\
			ailis@student.unimelb.edu.au, salehi.b@unimelb.edu.au, tb@ldwin.net, jianzhong.qi@unimelb.edu.au
		}
	\maketitle
	\begin{abstract}
		\begin{quote}

			The quality of a document is affected by various factors, including grammaticality, readability, stylistics, and expertise depth, making the task of document quality assessment a complex one. In this paper, we explore this task in the context of assessing the quality of Wikipedia articles and academic papers. Observing that the visual rendering of a document can capture implicit quality indicators that are not present in the document text --- such as images, font choices, and visual layout --- we propose a joint model that combines the text content with a visual rendering of the document for document quality assessment. Experimental results over two datasets reveal that textual and visual features are complementary, achieving state-of-the-art results.
		\end{quote}
	\end{abstract}
	
	\section{Introduction}
	
	The task of document quality assessment is to automatically assess a document according to some predefined inventory of quality labels.  This can take many forms, including essay scoring (quality = language quality, coherence, and relevance to a topic), job application filtering (quality = suitability for role + visual/presentational quality of the application), or answer selection in community question answering (quality = actionability + relevance of the answer to the question). In the case of this paper, we focus on document quality assessment in two contexts: Wikipedia document quality classification, and whether a paper submitted to a conference was accepted or not.
	
	\begin{figure}[!t]
          \centering
          \begin{subfigure}[t!]{0.45\textwidth}
            \begin{center}
              \includegraphics[width=\textwidth,height=3cm]{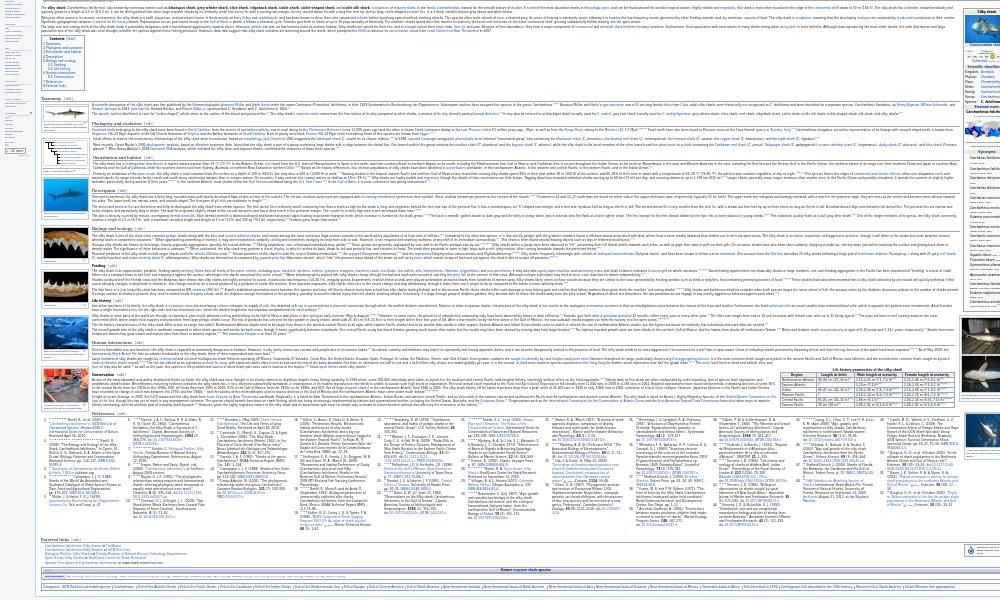}
            \end{center}
            \caption{Featured article}
            \label{FA_screenshots}
          \end{subfigure}
          \begin{subfigure}[t!]{0.45\textwidth}
            \begin{center}
            \includegraphics[width=\textwidth,height=3cm]{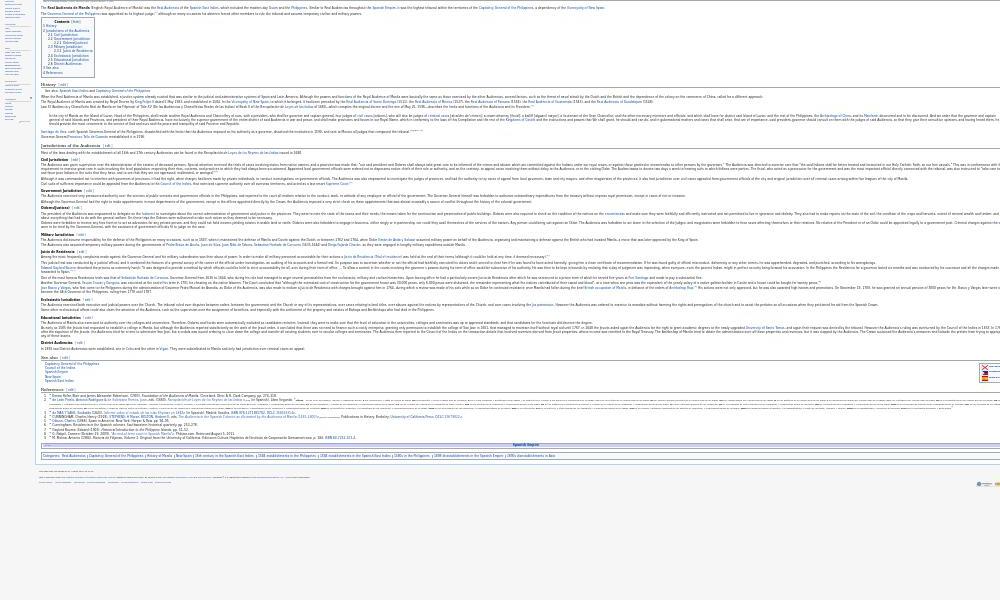}
          \end{center}
            \caption{Lower quality article}
            \label{C_screenshots}
          \end{subfigure}
          \caption{Visual renderings of two example Wikipedia documents
            with different quality labels (not intended to be readable).}
          \label{screenshots_examples}
	\end{figure}

	Automatic quality assessment has obvious benefits in terms of time savings and tractability in contexts where the volume of documents is large. In the case of dynamic documents (possibly with multiple authors), such as in the case of Wikipedia, it is particularly pertinent, as any edit potentially has implications for the quality label of that document (and around 10 English Wikipedia documents are edited per second\footnote{\url{https://en.wikipedia.org/wiki/Wikipedia:Statistics}}). Furthermore, when the quality assessment task is decentralized (as in the case of Wikipedia and academic paper assessment), quality criteria are often applied inconsistently by different people, where an automatic document quality assessment system could potentially reduce inconsistencies and enable immediate author feedback.

Current studies on document quality assessment mainly focus on textual features. For example, \citet{warncke2015success} examine features such as the article length and the number of headings to predict the quality class of a Wikipedia article. In contrast to these studies, in this paper, we propose to combine text features with visual features, based on a visual rendering of the document. \figref{screenshots_examples} illustrates our intuition, relative to Wikipedia articles. Without being able to read the text, we can tell that the article in \figref{FA_screenshots} has higher quality than \figref{C_screenshots}, as it has a detailed infobox, extensive references, and a variety of images. Based on this intuition, we aim to answer the following question: \textit{can we achieve better accuracy on document quality assessment by complementing textual features with visual features?}

Our visual model is based on fine-tuning an Inception V3 model \cite{Szegedy16} over visual renderings of documents, while our textual model is based on a hierarchical biLSTM. We further combine the two into a joint model. We perform experiments on two datasets: a Wikipedia dataset novel to this paper, and an arXiv dataset provided by \citet{kang18naacl} split into three sub-parts based on subject category. Experimental results on the visual renderings of documents show that implicit quality indicators, such as images and visual layout, can be captured by an image classifier, at a level comparable to a text classifier. When we combine the two models, we achieve state-of-the-art results over 3/4 of our datasets.

This paper makes the following contributions:
\begin{enumerate}[(i)]
	\item this is the first study to use visual renderings of documents to capture implicit quality indicators not present in the document text, such as document visual layout; experimental results show that we can obtain a 2.9\% higher accuracy using only visual renderings of documents compared with using only textual features over a Wikipedia dataset, and we can obtain competitive results over an arXiv dataset.%as state-of-the-art approaches on the arXiv dataset.   
	
	\item we further propose a joint model to predict document quality combining visual and textual features; we observe further improvements on the Wikipedia dataset and on two of the three arXiv subsets, indicating that visual and textual features are complementary.
	
	\item we construct a large-scale Wikipedia dataset with full textual data, visual renderings, and quality class labels; we also supplement the existing arXiv datasets with visual renderings of each document.
\end{enumerate}
All code and data associated with this research will be released on publication.

\section{Related Work}
\label{related_work}

A variety of approaches have been proposed for document quality assessment across different domains: Wikipedia article quality assessment, academic paper rating, content quality assessment in community question answering (cQA), and essay scoring. Among these approaches, some use hand-crafted features while others use neural networks to learn features from documents. For each domain, we first briefly describe feature-based approaches and then review neural network-based approaches. %At the end of this section, we will briefly describe models used in computer vision and the difference between our task and other tasks which utilize both images and text.

\textbf{Wikipedia article quality assessment}: Quality assessment of Wikipedia articles is a task that assigns a quality class label to a given Wikipedia article, mirroring the quality assessment process that the Wikipedia community carries out manually. Many approaches have been proposed that use features from the article itself, meta-data features (e.g., the editors, and Wikipedia article revision history), or a combination of the two. Article-internal features capture information such as whether an article is properly organized, with supporting evidence, and with appropriate terminology. For example, %\cite{blumenstock2008size} uses article length as a metric to assess the quality of Wikipedia articles. 
\citet{lipka2010identifying} use writing styles represented by binarized character trigram features to identify featured articles. \citet{warncke2013tell} and \citet{warncke2015success} explore the number of headings, images, and references in the article. \citet{dang2016measuring} use nine readability scores, such as the percentage of difficult words in the document, %Flesch reading-ease score \cite{kincaid1975derivation},
to measure the quality of the article. %Later, an online Objective Revision Evaluation Service \cite{halfaker2015artificial}, which is based on these last two studies, has been built to measure the quality of Wikipedia articles. 
Meta-data features, which are indirect indicators of article quality, are usually extracted from revision history, and the interaction between editors and articles. For example, one heuristic that has been proposed is that higher-quality articles have more edits \cite{Dalip17,Dalip14}.  %\cite{Dalip2009automatic,Dalip11,Dalip14,Dalip17}.
%\cite{Dondio07} and 
\citet{Wang11} use the percentage of registered editors and the total number of editors of an article. Article--editor dependencies have also been explored. For example, \citet{stein2007does} % and \citet{adler2008assigning}
use the authority of editors to measure the quality of Wikipedia articles, where the authority of editors is determined by the articles they edit. %\cite{Hu07}, \cite{Suzuki13}, and \cite{suzuki2015quality} take the review information (such as how many revisions the text survived) or the reviewers' reputation into consideration.

Deep learning approaches to predicting Wikipedia article quality have also been proposed. For example, \citet{dang2016quality} use a version of doc2vec \cite{LeM14} to represent articles, and feed the document embeddings into a four hidden layer neural network. \citet{shen2017hybrid} first obtain sentence representations by averaging words within a sentence, and then apply a biLSTM \cite{hochreiter1997long} to learn a document-level representation, which is combined with hand-crafted features as side information. \citet{Dang17a} exploit two stacked biLSTMs to learn document representations.

\textbf{Academic paper rating}: Academic paper rating is a relatively new task in NLP/AI, with the basic formulation being to automatically predict whether to accept or reject a paper. \citet{kang18naacl} explore hand-crafted features, such as the length of the title, whether specific words (such as \lex{outperform}, \lex{state-of-the-art}, and \lex{novel}) appear in the abstract, and an embedded representation of the abstract as input to different downstream learners, such as logistic regression, decision tree, and random forest. \citet{YangACL2018} exploit a modularized hierarchical convolutional neural network (CNN), where each paper section is treated as a module. For each paper section, they train an attention-based CNN, and an attentive pooling layer is applied to the concatenated representation of each section, which is then fed into a softmax layer.

\textbf{Content quality assessment in cQA}: Automatic quality assessment in cQA is the task of determining whether an answer is of high quality, selected as the best answer, or ranked higher than other answers. To measure answer content quality in cQA, researchers have exploited various features from different sources, such as the answer content itself, the answerer's profile, interactions among users, and usage of the content. The most common feature used is the answer length \cite{Jeon06,Suryanto09}, with other features including: syntactic and semantic features, such as readability scores. \cite{Agichtein08}; similarity between the question and the answer at lexical, syntactic, and semantic levels \cite{Agichtein08,Belinkov15,Hou15}; or user data (e.g., a user's status points or the number of answers written by the user). %Usage features, such as the number of clicks (views), are also beneficial in measuring content quality in cQA \cite{Burel12}. 
There have also been approaches using neural networks. %\cite{wu2016ecnu}, \cite{Hsu16}, and \cite{Suggu16a} leverage a bi-directional LSTM to learn high-level feature representations. 
For example, \citet{Suggu16a} combine CNN-learned representations with hand-crafted features to predict answer quality. \citet{Zhou15} use a 2-dimensional CNN to learn the semantic relevance of an answer to the question, and apply an LSTM to the answer sequence to model thread context. \citet{Guzman16} and \citet{Guzman16a} model the problem similarly to machine translation quality estimation, treating answers as competing translation hypotheses and the question as the reference translation, and apply neural machine translation to the problem. %With the assumption that similar answers should have similar quality labels, \cite{Joty15} and \cite{Joty16} predict answer quality at a thread level or in a joint way.

\textbf{Essay scoring}: Automated essay scoring is the task of assigning a score to an essay, usually in the context of assessing the language ability of a language learner. The quality of an essay is affected by the following four primary dimensions: topic relevance, organization and coherence, word usage and sentence complexity, and grammar and mechanics. To measure whether an essay is relevant to its ``prompt'' (the description of the essay topic), lexical and semantic overlap is commonly used \cite{Persing14,Phandi15}. %Lexical overlap and semantic similarity features are exploited to measure coherence between different discourse elements, sentences, and paragraphs \cite{Higgins04,mcnamara2015hierarchical}. 
\citet{attali2004automated} explore word features, such as the number of verb formation errors, average word frequency, and average word length, to measure word usage and lexical complexity. %Intelli-Metric \cite{rudner2006evaluation}, \citet{Yannakoudakis11} and 
\citet{Cummins16} use sentence structure features to measure sentence variety. The effects of grammatical and mechanic errors on the quality of an essay are measured via word and part-of-speech $n$-gram features and ``mechanics'' features \cite{Persing13} (e.g., spelling, capitalization, and punctuation), respectively. \citet{Taghipour16}, \citet{Alikaniotis16}, and \citet{TayPTH18} use an LSTM to obtain an essay representation, which is used as the basis for classification. Similarly, \citet{Dong17} utilize a CNN to obtain sentence representation and an LSTM to obtain essay representation, with an attention layer at both the sentence and essay levels.

\section{The Proposed Joint Model}
\label{model}

We treat document quality assessment as a classification problem, i.e., given a document, we predict its quality class (e.g., whether an academic paper should be accepted or rejected). The proposed model is a joint model that integrates visual features learned through Inception V3 with textual features learned through a biLSTM. In this section, we present the details of the visual and textual embeddings, and finally describe how we combine the two. We return to discuss hyper-parameter settings and the experimental configuration in the Experiments section.

\subsection{Visual Embedding Learning}

A wide range of models have been proposed to tackle the image classification task, such as VGG \cite{SimonyanZ14a}, ResNet \cite{He16}, Inception V3 \cite{Szegedy16}, and Xception \cite{Chollet17}. However, to the best of our knowledge, there is no existing work that has proposed to use visual renderings of documents to assess document quality. In this paper, we use Inception V3 pretrained on ImageNet\footnote{\url{http://www.image-net.org/}} (``\inception'' hereafter) to obtain visual embeddings of documents, noting that any image classifier could be applied to our task. The input to \inception is a visual rendering (screenshot) of a document, and the output is a visual embedding, which we will later integrate with our textual embedding.

Based on the observation that it is difficult to decide what types of convolution to apply to each layer (such as 3$\times$3 or 5$\times$5), the basic Inception model applies multiple convolution filters in parallel and concatenates the resulting features, which are fed into the next layer. This has the benefit of capturing both local features through smaller convolutions and abstracted features through larger convolutions. \inception is a hybrid of multiple Inception models of different architectures. To reduce computational cost, \inception also modifies the basic model by applying a 1$\times$1 convolution to the input and factorizing larger convolutions into smaller ones.

\subsection{Textual Embedding Learning}

We adopt a bi-directional LSTM model to generate textual embeddings for document quality assessment, following the method of \citet{shen2017hybrid} (``\bilstm'' hereafter). The input to \bilstm is a textual document, and the output is a textual embedding, which will later integrate with the visual embedding.

For \bilstm, each word is represented as a word embedding \cite{bengio2003neural}, and an average-pooling layer is applied to the word embeddings to obtain the sentence embedding, which is fed into a bi-directional LSTM to generate the document embedding from the sentence embeddings. Then a max-pooling layer is applied to select the most salient features from the component sentences.

\subsection{The Joint Model}

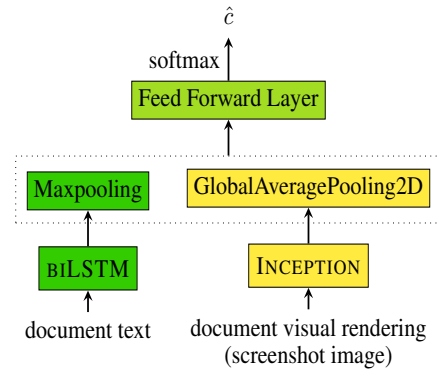
\begin{figure}[t]
	\begin{center}
		\scalebox{0.7}[0.8]{
			\resizebox{\columnwidth}{!}{
				\begin{tikzpicture}

%% Char RNN
%\node[align=center] at (0.0,0.0) (c_w_i_1) {$\vec{s}_{1}$};
%\node[align=center] at (1.5,0.0) (c_w_i_n) {$\vec{s}_{n}$};
\node[align=center] at (1.5,0.0) (s_0) {document text\\};
\node[align=center] at (5,0.0) (s_1) {document visual rendering\\ (screenshot image)};
\node[draw,align=center,minimum height=0.6cm, fill={rgb:green,2; red,0.5}] (bi_lstm) [above=0.3of s_0] {\bilstm};
\node[draw,align=center,minimum height=0.6cm, fill={rgb:orange,0;yellow,2;pink,1}] (inception) [above=0.3of s_1] {\inception};
\path[draw,thick,->,>=stealth] (s_0) to (bi_lstm);
\path[draw,thick,->,>=stealth] (s_1) to (inception);
\node[align=center,draw,inner sep=3, fill={rgb:orange,0;yellow,2;pink,1}] [above=0.5 of inception] (average) {GlobalAveragePooling2D};

\node[align=center,draw,inner sep=3, fill={rgb:green,2; red,0.5}] [above=0.5 of bi_lstm] (maxpooling) {Maxpooling};

\path[draw,thick,->,>=stealth] (bi_lstm) to (maxpooling);
\path[draw,thick,->,>=stealth] (inception) to (average);

\node[draw,dotted,fit=(maxpooling) (average),inner sep=5] (merge) {};

\node[align=center,draw,inner sep=3, fill={rgb:green,2; red,0.5;orange,0;yellow,2;pink,1}] [above=0.5 of merge] (ff) {Feed Forward Layer};

\path[draw,thick,->,>=stealth] (merge) to (ff);

\node[align=center,inner sep=5] [above=0.6 of ff] (y_pred) {$\hat{c}$};

\path[thick,->,>=stealth] (ff.north) edge node[anchor=center, left, midway] {softmax} (y_pred.south);

\end{tikzpicture}
			}
		}
	\end{center}
	\caption{\label{joint}Overview of the proposed model.}
\end{figure}

The proposed joint model (``\joint'' hereafter) combines the visual and textual embeddings (output of \inception and \bilstm) via a simple feed-forward layer and softmax over the document label set, as shown in \figref{joint}. We optimize our model based on cross-entropy loss.

\section{Experiments}
\label{experiment}
In this section, we first describe the two datasets used in our experiments: (1) \wikipedia, and (2) \arxiv. Then, we report the experimental details and results.

\subsection{Datasets}
\label{experiment_dataset}

\subsubsection{Wikipedia dataset}

The \wikipedia dataset consists of articles from English Wikipedia, with quality class labels assigned by the Wikipedia community. 
Wikipedia articles are labelled with one of six quality classes, in descending order of quality: {Featured Article} (``\FA''), {Good Article} (``\GA''), {B-class Article} (``\Bart''), {C-class Article} (``\Cart''), {Start Article} (``\Start''), and {Stub Article} (``\Stub''). A description of the criteria associated with the different classes can be found in the Wikipedia grading scheme page.\footnote{\url{https://en.wikipedia.org/wiki/Template:Grading_scheme}} The quality class of a Wikipedia article is assigned by Wikipedia reviewers or any registered user, who can discuss through the article's talk page\footnote{\url{https://en.wikipedia.org/wiki/Help:Talk_pages}} to reach consensus. We constructed the dataset by first crawling all articles from each quality class repository, e.g., we get \FA articles by crawling pages from the \FA repository: \url{https://en.wikipedia.org/wiki/Category:Featured_articles}.  
%After crawling from each quality category, we get 5278 \FA, 27543 \GA, 212146 \Bart, 533174 \Cart, 2636859 \Start, and 3214774 \Stub. 
This resulted in around 5K \FA, 28K \GA, 212K \Bart, 533K \Cart, 2.6M \Start, and 3.2M \Stub articles. 

We randomly sampled 5,000 articles from each quality class and removed all redirect pages, resulting in a dataset of 29,794 articles. As the wikitext contained in each document contains markup relating to the document category such as \lex{\{Featured Article\}} or \lex{\{geo-stub\}}, which reveals the label, we remove such information. We additionally randomly partitioned this dataset into training, development, and test splits based on a ratio of 8:1:1. Details of the dataset are summarized in \tabref{Wikipedia_data_distribution}.

We generate a visual representation of each document via a 1,000$\times$2,000-pixel screenshot of the article via a PhantomJS script over the rendered version of the article,\footnote{\url{https://github.com/ariya/phantomjs/blob/master/examples/rasterize.js}} ensuring that the screenshot and wikitext versions of the article are the same version. Any direct indicators of document quality (such as the \FA indicator, which is a bronze star icon in the top right corner of the webpage) are removed from the screenshot.
%modified version of webscreenshot \footnote{\url{https://github.com/maaaaz/webscreenshot}}. 

\begin{table}[t]  
	
	\scalebox{1.0}{
		\begin{tabular}{lcccc}  
			\toprule  
			{Class} &{Train}       &{Dev}       &{Test} &{Total}            \\  \midrule
			\FA     &4000      &500      &500  &5000\\
			\GA     &4000      &500      &500  &5000\\
			\Bart   &4000      &500      &455  &4955\\
			\Cart   &4000      &500      &467  &4967\\
			\Start  &4000      &500      &451  &4951\\
			\Stub   &4000      &500      &421  &4921\\ \midrule
			Total  &24000\z     &3000\z     &2794\z  &29794\z
			\\ \bottomrule
	\end{tabular}  }
	\centering 
	\caption{\wikipedia dataset.}  
	\label{Wikipedia_data_distribution}  
\end{table}

\subsubsection{\arxiv dataset}

The \arxiv dataset \cite{kang18naacl} consists of three subsets of
academic articles under the arXiv repository of Computer Science (cs),
from the three subject areas of: Artificial Intelligence (\csai),
Computation and Language (\cscl), and Machine Learning (\cslg). In line
with the original dataset formulation \cite{kang18naacl}, a paper is
considered to have been accepted (i.e.\ is positively labeled) if it
matches a paper in the DBLP database or is otherwise accepted by any of
the following conferences: ACL, EMNLP, NAACL, EACL, TACL, NIPS, ICML, ICLR, or AAAI. Failing this, it is considered to be rejected (noting that some of the papers may not have been submitted to one of these conferences). The median numbers of pages for papers in cs.ai, cs.cl, and cs.lg are 11, 10, and 12, respectively. To make sure each page in the PDF file has the same size in the screenshot, we crop the PDF file of a paper to the first 12; we pad the PDF file with blank pages if a PDF file has less than 12 pages, using the PyPDF2 Python package.\footnote{\url{https://pypi.org/project/PyPDF2/}} We then use ImageMagick\footnote{\url{https://www.imagemagick.org/script/index.php}} to convert the 12-page PDF file to a single 1,000$\times$2,000 pixel screenshot. \tabref{arxiv_data_distribution} details this dataset, where the ``Accepted'' column denotes the percentage of positive instances (accepted papers) in each subset.

\begin{table}[t]  	
  \centering 
  \begin{tabular}{lccccc}  
    \toprule  
    Subject  & Accepted   &{Train}       &{Dev}       &{Test} &{Total}            \\  \midrule
    \csai     &10\%   &3682      &205      &205  &4092\\
    \cscl     &30\%   &2374      &132      &132  &2638\\
    \cslg     &32\%   &4543      &252      &253  &5048\\ \bottomrule
  \end{tabular}
  \caption{\arxiv dataset. ``Accepted'' indicates the proportion of accepted papers in the given subject.}  
  \label{arxiv_data_distribution}  
\end{table}

\subsection{Experimental Setting}
%For the document content, we divide an article into sentences and tokenize them using \emph{NLTK} \cite{bird2006nltk}. Words appearing more than 20 times are retained when building the vocabulary. All other words are replaced by the special \texttt{UNK} token. We use the pre-trained \textit{GloVe} \cite{pennington2014glove} 50-dimensional word embeddings to represent words. For words that cannot be found in GloVe, word embeddings are randomly initialized based on sampling from a uniform distribution $U(-1, 1)$. All word embeddings are updated in the training process. 

As discussed above, our model has two main components --- \bilstm and \inception --- which generate textual and visual representations, respectively. For the \bilstm component, the documents are preprocessed as described in \citet{shen2017hybrid}, where an article is divided into sentences and tokenized using {NLTK} \cite{bird2006nltk}. Words appearing more than 20 times are retained when building the vocabulary. All other words are replaced by the special \texttt{UNK} token. We use the pre-trained {GloVe} \cite{pennington2014glove} 50-dimensional word embeddings to represent words. For words not in GloVe, word embeddings are randomly initialized based on sampling from a uniform distribution $U(-1, 1)$. All word embeddings are updated in the training process. We set the LSTM hidden layer size to 256. The concatenation of the forward and backward LSTMs thus gives us 512 dimensions for the document embedding. A dropout layer is applied at the sentence and document level, respectively, with a probability of 0.5.

For \inception, we adopt data augmentation techniques in the training with  a ``nearest'' filling mode, a zoom range of 0.1, a width shift range of 0.1, and a height shift range of 0.1. As the original screenshots have the size of 1,000$\times2$,000 pixels, they are resized to 500$\times$500 to feed into \inception, where the input shape is (500, 500, 3). A dropout layer is applied with a probability of 0.5. Then, a GlobalAveragePooling2D layer is applied, which produces a 2,048 dimensional representation.  

For the \joint model, we get a representation of 2,560 dimensions by concatenating the 512 dimensional representation from the \bilstm with the 2,048 dimensional representation from \inception. The dropout layer is applied to the two components with a probability of 0.5. For \bilstm, we use a mini-batch size of 128 and a learning rate of 0.001. For both \inception and joint model, we use a mini-batch size of 16 and a learning rate of 0.0001. All hyper-parameters were set empirically over the development data, and the models were optimized using the Adam optimizer \cite{adam2014}. 

In the training phase, the weights in \inception are initialized by parameters pretrained on ImageNet, and the weights in \bilstm are randomly initialized (except for the word embeddings). We train each model for 50 epochs. However, to prevent overfitting, we adopt early stopping, where we stop training the model if the performance on the development set does not improve for 20 epochs. For evaluation, we use (micro-)accuracy, following previous studies \cite{dang2016measuring,kang18naacl}. 

\subsection{Baseline Approaches}

We compare our models against the following five baselines: 
\begin{itemize}
\item \majority: the model labels all test samples with the majority class of the training data. 
\item \benchmark: a benchmark method from the literature. In the case of \wikipedia, this is \citet{dang2016measuring}, who use structural features and readability scores as features to build a random forest classifier; for \arxiv, this is \citet{kang18naacl}, who use hand-crafted features, such as the number of references and TF-IDF weighted bag-of-words in abstract, to build a classifier based on the best of logistic regression, multi-layer perception, and AdaBoost.
\item \doctovec: doc2vec \cite{LeM14} to learn document embeddings with
  a dimension of 500, and a 4-layer feed-forward classification model on
  top of this, with 2000, 1000, 500, and 200 dimensions, respectively.
\item \bilstm: first derive a sentence representation by averaging across words in a sentence, then feed the sentence representation into a biLSTM and a maxpooling layer over output sequence to learn a document level representation with a dimension of 512, which is used to predict document quality.
\item \frozen: the frozen \inception model, where only parameters in the last layer are fine-tuned during training.
\end{itemize}
The hyper-parameters of \benchmark, \doctovec, and \bilstm are based on the corresponding papers
except that: (1) we fine-tune the feed forward layer of \doctovec on the development set and train the model 300 epochs on \wikipedia and 50 epochs on \arxiv; (2) we do not use hand-crafted features for \bilstm as we want the baselines to be comparable to our models, and the main focus of this paper is not to explore the effects of hand-crafted features (e.g., see \citet{shen2017hybrid}).

\begin{table*}[!t]  
	\centering   
	\scalebox{0.94}{
		\begin{threeparttable}
			\begin{tabular}{lcccccccc}  
				\toprule  
				%\diagbox{\multicolumn{2}{l}{accuracy}}{\multicolumn{2}{l}{accuracy}}
				%\multicolumn{2}{l}{accuracy} 
                          
                          &&\majority &\benchmark      &\doctovec  &\frozen &\bilstm  &\inception  &\joint      \\    \midrule 
				\multicolumn{2}{l}{\wikipedia} & 16.7\%       &46.7$\pm$0.34\% &23.2$\pm$1.41\%  & 43.7$\pm$0.51 &54.1$\pm$0.47\%  &57.0$\pm$0.63\%   &\textbf{59.4$\pm$0.47\%$^{\dagger}$}  \\ \midrule %   
				
				\multirow{3}{*}{\arxiv} 
				&\csai    & 92.2\%  & 92.6\%  & 73.3$\pm$9.81\%  &92.3$\pm$0.29  & 91.5$\pm$1.03\%   & 92.8$\pm$0.79\%  & \textbf{93.4$\pm$1.07\%$^{\ssig}$} \\
				&\cscl    & 68.9\%  & 75.7\%  & 66.2$\pm$8.38\% & 75.0$\pm$1.95   &76.2$\pm$1.30\%    & 76.2$\pm$2.92\%  &\textbf{77.1$\pm$3.10\%$^{\nosig}$}  \\
				&\cslg    & 67.9\%  & 70.7\%  & 64.7$\pm$9.08\% & 73.9$\pm$1.23 &\textbf{81.1$\pm$0.83\%}    & 79.3$\pm$2.94\%  & 79.9$\pm$2.54\%$^{\nosig}$  \\
				\bottomrule
			\end{tabular}  %}
			
	\end{threeparttable}}
      \caption{Experimental results. The best result for each dataset is indicated in \textbf{bold}, and marked with ``\ssig'' if it is significantly higher than the second best result (based on a one-tailed Wilcoxon signed-rank test; $p<0.05$).
        The results of \benchmark on the \arxiv dataset are from the original paper, where the standard deviation values were not reported. All neural models except for \frozen have larger standard deviation values on \arxiv than \wikipedia, which can be explained by the small size of the \arxiv test set.}  
	\label{comparisons}  
\end{table*} 

\subsection{Experimental Results}

\tabref{comparisons} shows the performance of the different models over our two datasets, in the form of the average accuracy on the test set (along with the standard deviation) over 10 runs, with different random initializations.    

On \wikipedia, we observe that the performance of \bilstm, \inception, and \joint is much better than that of all four baselines. %\textsf{Doc2Vec} achieves an accuracy of $45.7\%$, which is still much worse than \bilstm, even if the test set is included to train the \textsf{Doc2Vec} to get the document embeddings. 
\inception achieves 2.9\% higher accuracy than \bilstm. The performance of \joint achieves an accuracy of 59.4\%, which is 5.3\% higher than using textual features alone (\bilstm) and 2.4\% higher than using visual features alone (\inception). Based on a one-tailed Wilcoxon signed-rank test, the performance of \joint is statistically significant ($p<0.05$). This shows that the textual and visual features complement each other, achieving state-of-the-art results in combination.

For \arxiv, baseline methods \majority, \benchmark, and \frozen outperform \bilstm over \csai, in large part because of the class imbalance in this dataset (90\% of papers are rejected). Surprisingly, \frozen is better than \majority and \benchmark over the arXiv \cslg subset, which verifies the usefulness of visual features, even when only the last layer is fine-tuned. \tabref{comparisons} also shows that \inception and \bilstm achieve similar performance on \arxiv, showing that textual and visual representations are equally discriminative: \inception and \bilstm are indistinguishable over \cscl; \bilstm achieves 1.8\% higher accuracy over \cslg, while \inception achieves 1.3\% higher accuracy over \csai. Once again, the \joint model achieves the highest accuracy on \csai and \cscl by combining textual and visual representations (at a level of statistical significance for \csai). This, again, confirms that textual and visual features complement each other, and together they achieve state-of-the-art results. On \arxiv \cslg, \joint achieves a 0.6\% higher accuracy than \inception by combining visual features and textual features, but \bilstm achieves the highest accuracy. One characteristic of \cslg documents is that they tend to contain more equations than the other two \arxiv datasets, and preliminary analysis suggests that the \bilstm is picking up on a correlation between the volume/style of mathematical presentation and the quality of the document.
% TJB: Aili, would you agree with this?

\section{Analysis}
\label{analysis}

In this section, we first analyze the performance of \inception and \joint. We also analyze the performance of different models on different quality classes. The high-level representations learned by different models are also visualized and discussed. As the \wikipedia test set is larger and more balanced than that of \arxiv, our analysis will focus on \wikipedia.

\subsection{\inception}
\label{Inception_analysis}

\begin{figure}[!t]
	\centering
        \begin{subfigure}[t!]{0.45\textwidth}
          \begin{center}
            \includegraphics[width=0.7\linewidth]{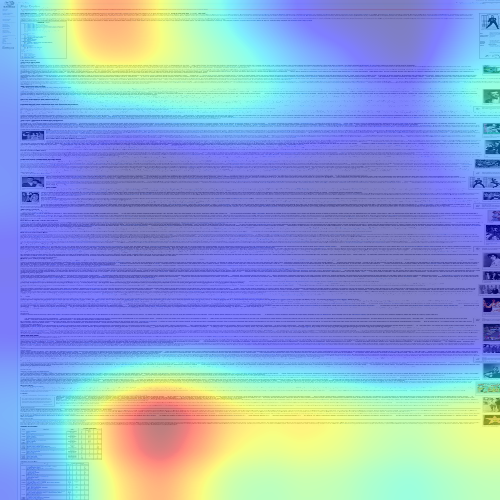}
          \end{center}
          \caption{\FA}
          \label{FA_image1}
        \end{subfigure}
        \begin{subfigure}[t!]{0.45\textwidth}
          \begin{center}
            \includegraphics[width=0.7\linewidth]{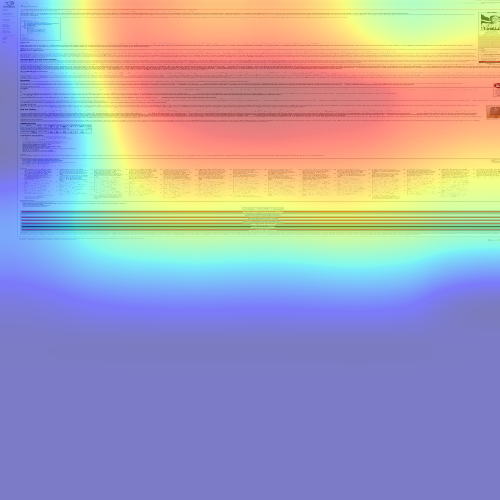}
          \end{center}
          \caption{\GA}
          \label{GA_image1}
        \end{subfigure}

        \vspace*{3mm}

        \begin{subfigure}[t!]{0.45\textwidth}
          \begin{center}
            \includegraphics[width=0.7\linewidth]{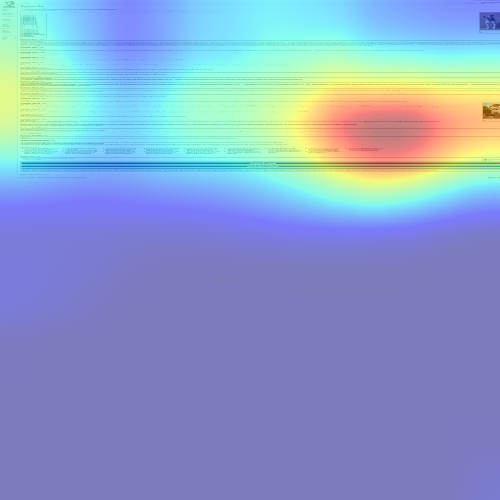}
          \end{center}
          \caption{\Bart}
          \label{B_image1}
        \end{subfigure}
        \begin{subfigure}[t!]{0.45\textwidth}
          \begin{center}
            \includegraphics[width=0.7\linewidth]{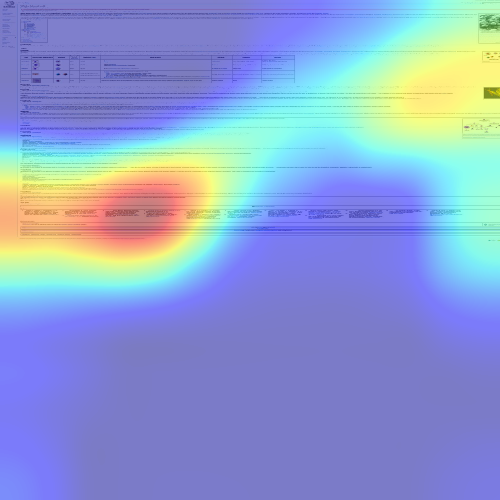}
          \end{center}
          \caption{\Cart}
          \label{C_image1}
        \end{subfigure}
        
        \vspace*{3mm}

        \begin{subfigure}[t!]{0.45\textwidth}
          \begin{center}
            \includegraphics[width=0.7\linewidth]{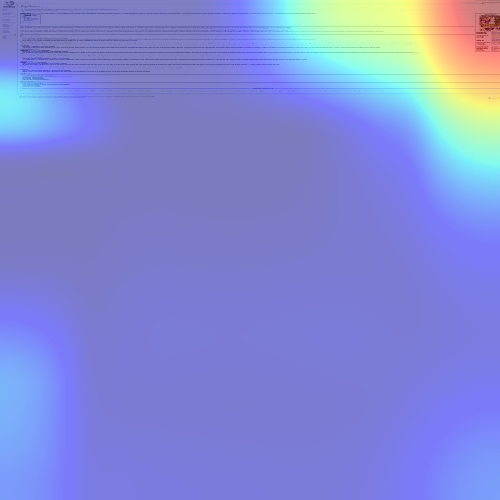}
          \end{center}
          \caption{\Start}
          \label{Start_image1}
        \end{subfigure}
        \begin{subfigure}[t!]{0.45\textwidth}
          \begin{center}
            \includegraphics[width=0.7\linewidth]{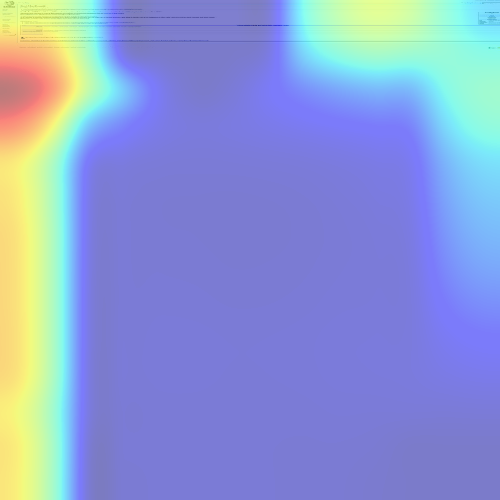}
          \end{center}
          \caption{\Stub}
          \label{Stub_image1}
        \end{subfigure}

        \caption{Heatmap overlapped onto screenshots of each Wikipedia quality class. Best viewed in color.}
		\label{class_activation_map}
\end{figure}

%	\begin{figure}[!t]
%		\centering
%		\scalebox{0.85}{
%			\begin{tabularx}{\linewidth}{@{}cXX@{}}
%				\begin{tabular}{c}
%					\subfloat[\FA\label{FA_image2}]{%
%						\includegraphics[width=0.3\linewidth]{./saliency/FA_4005_heatmap_saliency.png}
%					} 
%					\\
%					\subfloat[\GA\label{GA_image2}]{%
%						\includegraphics[width=0.3\linewidth]{./saliency/GA_9064_heatmap_saliency.png}
%					}
%				\end{tabular}
%				
%				\begin{tabular}{c}
%					\subfloat[\Bart\label{B_image2}]{%
%						\includegraphics[width=0.3\linewidth]{./saliency/B_14001_heatmap_saliency.png}
%					} 
%					\\
%					\subfloat[\Cart\label{C_image2}]{%
%						\includegraphics[width=0.3\linewidth]{./saliency/C_18965_heatmap_saliency.png}
%					}
%				\end{tabular}
%				
%				\begin{tabular}{c}
%					\subfloat[\Start\label{Start_image2}]{%
%						\includegraphics[width=0.3\linewidth]{./saliency/Start_23923_heatmap_saliency.png}
%					}
%					\\
%					\subfloat[\Stub\label{Stub_image2}]{%
%						\includegraphics[width=0.3\linewidth]{./saliency/Stub_28874_heatmap_saliency.png}
%					}
%				\end{tabular}
%				
%			\end{tabularx}
%			\caption{Saliency map of Wikipedia quality classes. Best viewed in color.} 
%			\label{attention_heatmap_saliency}
%		}
%		
%	\end{figure}

To better understand the performance of \inception, we generated the gradient-based class activation map \cite{Selvaraju17}, by maximizing the outputs of each class in the penultimate layer, as shown in 
\figref{class_activation_map}. From \figref{FA_image1} and \figref{GA_image1}, we can see that \inception identifies the two most important regions (one at the top corresponding to the table of contents, and the other at the bottom, capturing both document length and references) that contribute to the \FA class prediction, and a region in the upper half of the image that contributes to the \GA class prediction (capturing the length of the article body). From \figref{B_image1} and \figref{C_image1}, we can see that the most important regions in terms of \Bart and \Cart class prediction capture images (down the left and right of the page, in the case of \Bart and \Cart), and document length/references. From \figref{Start_image1} and \figref{Stub_image1}, we can see that \inception finds that images in the top right corner are the strongest predictor of \Start class prediction, and (the lack of) images/the link bar down the left side of the document are the most important for \Stub class prediction. %The class saliency map \cite{Simonyan13} show that \inception can capture information such as article length. %\figref{attention_heatmap_saliency} is the class saliency map \cite{Simonyan13}, which indicates pixels whose change would affect the class score the most. From \figref{attention_heatmap_saliency} $(a)$ to \figref{attention_heatmap_saliency} $(f)$, we can see that the saliency heatmap can capture the information such as article length.

\subsection{\joint}
\begin{table}[t]  
	\centering  
		\begin{tabular}{lcccccc}  
			\toprule  
			
			Quality &\FA    &\GA    &\Bart        &\Cart      &\Start    &\Stub   \\  \midrule 
			\FA  &\cellcolor{gray!30}397  &83    &20       &0      &0        &0        \\
			\GA  &112   &\cellcolor{gray!30}299   &65      &22     &2        &0        \\
			\Bart   &23    &53    &\cellcolor{gray!30}253      &75    &44       &7      \\
			\Cart   &5    &33    &193      &\cellcolor{gray!30}124    &100       &12   \\
			\Start   &1     &6     &36       &85    &\cellcolor{gray!30}239      &84     \\
			\Stub    &0     &0     &6        &7     &63      &\cellcolor{gray!30}345     \\  \bottomrule
			
	\end{tabular}   
	\caption{Confusion matrix of the \joint model on \wikipedia. Rows are the actual quality classes and columns are the predicted quality classes. The diagonal (gray cells) indicates correct predictions. }  
	\label{confusion_matrix}  
\end{table}

%	\begin{table*}[tb]  
%		\centering  
%		\scalebox{1.0}{
%			\begin{tabular}{lcccccccc}  
%				\toprule  
%
%				Quality &\FA    &\GA    &\Bart        &\Cart      &\Start    &\Stub    &Class Total  &Accuracy  \\  \midrule 
%				\FA  &\cellcolor{gray!30}397  &83    &20       &0      &0        &0       &500  &79.4\% \\
%				\GA  &112   &\cellcolor{gray!30}299   &65      &22     &2        &0       &500  &59.8\% \\
%				\Bart   &23    &53    &\cellcolor{gray!30}253      &75    &44       &7      &455  &55.6\% \\
%				\Cart   &5    &33    &193      &\cellcolor{gray!30}124    &100       &12      &467  &26.6\% \\
%				\Start   &1     &6     &36       &85    &\cellcolor{gray!30}239      &84     &451  &53.0\% \\
%				\Stub    &0     &0     &6        &7     &63      &\cellcolor{gray!30}345     &421  &81.9\%\\  \midrule
%				Predicted Total   &538  &474   &573     &313   &448      &448   &2794 &69.36\% 
%					\\
%				Precision  &73.8\%   &63.1\%  &44.2\% &39.6\% &53.3\% &77.0\%
%				\\ \bottomrule
%			\end{tabular}   }
%		\caption{Confusion matrix of the \joint model on the Wikipedia test set. Rows are actual quality classes, and columns are the predicted quality classes. The last column is the accuracy for each class, and the last row is the precision for each class. Diagonal elements (gray cells) of the matrix are correct predictions. }  
%		\label{confusion_matrix}  
%		%Here, \FA stands for featured article; \GA stands for good article; \Bart stands for B-class article; \Cart stands for C-class article; \Start stands for start article; and \Stub stands for stub article.
%\end{table*} 

\begin{table}[!t]  
	\centering 
		\begin{tabular}{lcccc}%{lp{12mm}p{12mm}p{13mm}p{12mm}}
			\toprule
			Quality &Metric &\bilstm  &\inception &\joint \\ \hline
			\multirow{3}{*}{\FA} & \precision &\textbf{76.6} & 74.8 & 73.8 \\
			&\recall & 72.0 & 68.2 & \textbf{79.4} \\
			&\fscore     &74.2  &71.3  & \textbf{76.5} \\ \midrule
			
			\multirow{3}{*}{\GA} & \precision &51.3 & 57.7 & \textbf{63.1} \\
			&\recall & \textbf{59.8} & 59.0 & \textbf{59.8} \\
			&\fscore     &55.2  & 58.3  & \textbf{61.4} \\ \midrule
			
			\multirow{3}{*}{\Bart} & \precision &37.6 & 41.8 & \textbf{44.2} \\
			&\recall & 42.4 & 44.0 & \textbf{55.6} \\
			&\fscore     &39.9  &42.9  & \textbf{49.2} \\ \midrule
			
			\multirow{3}{*}{\Cart} & \precision &36.3 & 38.9 & \textbf{39.6} \\
			&\recall & 27.0 & \textbf{36.0} & 26.6 \\
			&\fscore     &31.0  &\textbf{37.4}  & 31.8 \\ \midrule
			
			\multirow{3}{*}{\Start} & \precision &48.2 & 49.4 & \textbf{53.3} \\
			&\recall & 44.8 & \textbf{57.2} & 53.0 \\
			&\fscore     &46.4  &53.0  & \textbf{53.1} \\ \midrule
			
			\multirow{3}{*}{\Stub} & \precision &71.9 & \textbf{83.3} & 77.0 \\
			&\recall & 78.9 & 78.2 & \textbf{81.9} \\
			&\fscore     &75.2  &\textbf{80.7}  & 79.4 \\  \bottomrule
		\end{tabular}
	\caption{Precision (``\precision''), recall (``\recall''), and F1 (``\fscore'') of \bilstm, \inception, and \joint on \wikipedia.}  
	\label{precision_recall_f1}  
\end{table}

\begin{figure*}[t!]
	\begin{center}
		\scalebox{0.8}{
			\resizebox{\columnwidth}{!}{
				\input{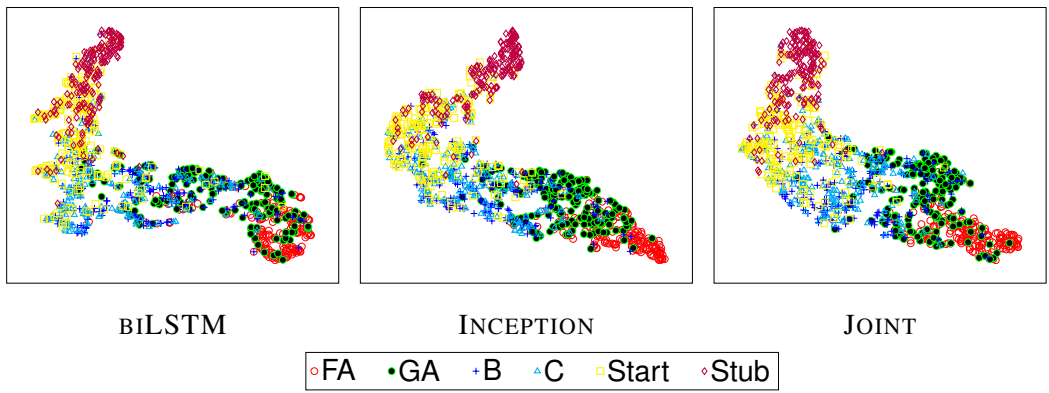}
			}
		}
	\end{center}
	\caption{t-SNE scatter plot of \wikipedia article representations (representations from penultimate layer of each model, 200 random samples from each quality class; best viewed in color)}
        \label{class_scatter_plot}
\end{figure*}

\tabref{confusion_matrix} shows the confusion matrix of \joint on \wikipedia. We can see that more than 50\% of documents for each quality class are correctly classified, except for the \Cart class where more documents are misclassified into \Bart. Analysis shows that when misclassified, documents are usually misclassified into adjacent quality classes, which can be explained by the Wikipedia grading scheme, where the criteria for adjacent quality classes are more similar.\footnote{Suggesting that ordinal regression should boost accuracy, but preliminary experiments with various methods led to no improvement over simple classification.}

We also provide a breakdown of precision (``\precision''), recall (``\recall''), and F1 score (``\fscore'') for \bilstm, \inception, and \joint across the quality classes in \tabref{precision_recall_f1}. We can see that \joint achieves the highest accuracy in 11 out of 18 cases. It is also worth noting that all models achieve higher scores for \FA, \GA, and \Stub articles than \Bart, \Cart and \Start articles. This can be explained in part by the fact that \FA and \GA articles must pass an official review based on structured criteria, and in part by the fact that \Stub articles are usually very short, which is discriminative for \inception, and \joint. All models perform worst on the \Bart and \Cart quality classes. It is difficult to differentiate \Bart articles from \Cart articles even for Wikipedia contributors. As evidence of this, when we crawled a new dataset including talk pages with quality class votes from Wikipedia contributors, we found that among articles with three or more quality labels, over 20\% percent of \Bart and \Cart articles have inconsistent votes from Wikipedia contributors, whereas for \FA and \GA articles the number is only 0.7\%. 

We further visualize the learned document representations of \bilstm, \inception, and \joint in the form of a t-SNE plot \cite{vdMaaten:Hinton:2008} in \figref{class_scatter_plot}. The degree of separation between \Start and \Stub achieved by \inception is much greater than for \bilstm, with the separation between \Start and \Stub achieved by \joint being the clearest among the three models. \inception and \joint are better than \bilstm at separating \Start and \Cart. \joint achieves slightly better performance than \inception in separating \GA and \FA. We can also see that it is difficult for all models to separate \Bart and \Cart, which is consistent with the findings of \tabref[s]{confusion_matrix} and \ref{precision_recall_f1}.

\section{Conclusions}
\label{conclusions}

We proposed to use visual renderings of documents to capture implicit document quality indicators, such as font choices, images, and visual layout, which are not captured in textual content. We applied neural network models to capture visual features given visual renderings of documents. Experimental results show that we achieve a 2.9\% higher accuracy than state-of-the-art approaches based on textual features over \wikipedia, and performance competitive with or surpassing state-of-the-art approaches over \arxiv. We further proposed a joint model, combining  textual and visual representations, to predict the quality of a document. Experimental results show that our joint model outperforms the visual-only model in all cases, and the text-only model on \wikipedia and two subsets of \arxiv. These results underline the feasibility of assessing document quality via visual features, and the complementarity of visual and textual document representations for quality assessment.

\bibliography{aaai_simplified.bib}	
\bibliographystyle{aaai}
\end{document}